# Evolution of artificial intelligence languages – a systematic literature review

Emmanuel Adetiba[1,2], Temitope M. John[1], Adekunle A. Akinrinmade[1], Funmilayo S. Moninuola[1], Oladipupo O. Akintade[1], Joke A. Badejo[1]

[1]Department of Electrical and Information Engineering, College of Engineering, Covenant University, Ota, Nigeria.

[2]HRA, Institute for Systems Science, Durban University of Technology, P.O. Box 1334, Durban, South Africa.

Corresponding email: emmanuel.adetiba@covenantuniversity.edu.ng

## Abstract

The field of Artificial Intelligence (AI) has undoubtedly received significant attention in recent years. AI is being adopted to provide solutions to problems in fields such as medicine, engineering, education, government and several other domains. In order to analyze the state of the art of research in the field of AI, we present a systematic literature review focusing on the Evolution of AI programming languages. We followed the systematic literature review method by searching relevant databases like SCOPUS, IEEE Xplore and Google Scholar. EndNote reference manager was used to catalog the relevant extracted papers. Our search returned a total of 6565 documents, whereof 69 studies were retained. Of the 69 retained studies, 15 documents discussed LISP programming language, another 34 discussed PROLOG programming language, the remaining 20 documents were spread between Logic and Object Oriented Programming (LOOP), ARCHLOG, Epistemic Ontology Language with Constraints (EOLC), Python, C++, ADA and JAVA programming languages. This review provides information on the year of implementation, development team, capabilities, limitations and applications of each of the AI programming languages discussed. The information in this review could guide practitioners and researchers in AI to make the right choice of languages to implement their novel AI methods.

## Keywords

Artificial Intelligence; Programming Language; Python; AI; LISP; PROLOG; JAVA; C++; EOLC; ADA

## 1. Introduction

Artificial intelligence (AI) is concerned with intelligent behaviors in artifacts such as perception, reasoning, learning, communicating and acting in complex environment. AI is concerned about machines possessing the listed characteristics as well as humans can, or even better and faster. The physical symbol system hypothesis states that a physical symbol system has the necessary and sufficient means for general intelligent action. A physical symbol system is a machine like a digital computer that is capable of manipulating symbolic data- adding numbers, rearranging lists of symbols and replacing some symbols by others [1].

Traditionally, computing is used for performing mechanical computations using fixed procedures. This approach implies complex problems would be more difficult to solve. Another shortcoming is that computers so programmed would have difficulties understanding and adapting to new situations like human's do. AI is different from this traditional approach in that it requires machines to think and tackle such complex assignments. AI was formally coined by John McCarthy in a workshop conducted by IBM at Dartmouth College in 1956 [1].

When digital computers were first developed in the 1940s and 1950s researchers wrote a number of programs, these programs could play chess, checkers and prove theorems. In the 1960s and 1970s, AI explored various ways to represent problems by developing different search techniques and general heuristics, these enabled development of programs used to solve algebraic word problems and symbolic integration. In the 1970s and 1980s, as a result of more powerful systems, AI programs were used to build expert systems and by 1997 an IBM program named DEEP BLUE defeated the world chess champion, Garry Kasparov. Interest in AI sagged in the late 1950s, it, however, resumed with vigor in the 1980s. Networks of nonlinear elements with adjustable-strength interconnections are now recognized as an important class of non-linear modeling tools. There are now several important applications of AI [1]. AI programming languages are languages capable of implementation of logic [2], vastly useful for building expert systems [3] and with features for handling relational databases and natural language processing [4]. There are many AI programming languages amongst which are List processor (LISP), French for programming in logic (PROLOG), Python, JAVA, C++ and ADA.

LISP is a computer programming language with a long history and a distinctive, fully parenthesized prefix notation. [5] First conceived in 1958. It became the programming language of choice for AI research. It was the basis for many ideas in computer science, including tree data structures, automatic storage management, dynamic typing, conditionals, higher-order functions, recursion, the self-hosting compiler, [6] and the read–eval–print loop [7]. Linked lists are one of LISP's major data structures, and its LISP source code is made of lists. Thus, LISP programs can manipulate source code as a data structure, giving rise to the macro systems that allow programmers to create new syntax or new domain-specific languages embedded in LISP.

The idea of PROLOG was first conceived in 1970 and thereafter implemented in 1972 [8] by a group of AI scientists, namely, Alain Colmerauer , Robert Kowalski and Philippe Roussel. It was one of the most popular programming languages since its inception and remains the most popular with a lot of its variants [9]. Prolog is widely used for programming in AI being a general-purpose logic programming language [10] [11]. It differs from other programming languages in that it is intended mainly as a declarative programming language. It has its roots in first-order logic expressed in terms of relations, represented as facts and rules whilst a computation is initiated by running queries over these relations[12]. The language is well suited for different branches of AI, for example logical problems that are randomly selected[9] [13], expert systems [14] and natural language processing [15-16].

Python is a general-purpose, high-level programming language whose design philosophy emphasizes code readability. It was developed by Guido Van Rossum in the early 1990s[17]. Python's syntax allows programmers to express concepts in fewer lines of code than would be possible in languages such as C. [18]. Python is well adapted for AI tasks [19-20], especially in the area of natural language processing. It was conceived towards the end of 1980 [21] but its implementation began in December 1989 [22] as a successor to the ABC language capable of exception handling and interfacing with the Amoeba operating system [23]. By October 2000, Python 2.0 was released with new features which included a full garbage collector and support for Unicode [24], versions 2.6 and 2.7 followed while Python 3.0 was released on December 3, 2008 [25]. A number of Artificial Intelligence libraries run primarily on the Python Infrastructure; Keras, noted for its user friendliness, modularity and easy extensibility, is a high level neural network API compatible with Python 2.7-3.6[26]; Theano allows for the evaluation of mathematical expressions using multi-dimensional arrays efficiently [27] etc.

Java was developed in early 1990s by James Gosling from Sun Microsystems [28]. Some of the features of Java is that it can easily be coded, it is highly scalable, making it desirable for AI projects[29]. It is also portable, and can easily be implemented on different platforms since it uses virtual machine technology [30]. C++ was developed in 1979 by Bjarne Stroustrup at Bells Labs and standardized in 1998 as C language extension. Of all the AI programming language, it is the fastest and mostly used by developers in AI projects because of its time-sensitivity and also when speed is of higher priority to improve their project execution time [31].

We reviewed publicly available documents on AI programming languages from SCOPUS, IEEE Xplore and Google scholar. The documents were reviewed to retrieve the year of execution, development team, capabilities and features as well as the limitations and applications of these AI programming languages.

## 2. Methods

In order to provide a systematic review of AI programming languages, we followed the guidelines put forward by Kitchenham [32]. The goal of this systematic review is to access original articles and full length review articles that relates to AI programming languages. The process is detailed below:

### 2.1 Research Questions

The research questions address forthwith are:

RQ 1: What is the prevalence of AI programming language publications since 1963?

RQ 2: Which of the AI programming languages have received more attention with respect to the volume of research publication been produced?

RQ 3: What are the predominant AI programming languages on which recent AI softwares rely?

RQ 4: What are the characteristics of AI programming languages which makes them suitable or unsuitable for use across platforms?

In addressing RQ 1, we considered the volume of the publications published for each AI programming languages since 1963 each of these programming languages has been described in Table 2 - 11. With respect to RQ 2, we discussed the major AI programming languages; LISP, Prolog, LOOP, ARCHLOG, EOLC, Phyton, C++, ADA and JAVA. We considered the popularity of the languages with respect to the number of publications that have been produced from 1963 to 10 February 2018. To answer RQ 3; we analyze a few recently developed AI software platforms and determined the foundation AI architecture which oversees the operations of these new software.

With respect to the characteristics of AI programming languages (RQ 4), we considered the following; capabilities or features, limitations and application of the AI programming language.

## 2.2 Search Strategy

The search for relevant literature was conducted on the 10 February 2018 using SCOPUS and IEEE Xplore databases, the dates was set to 1963 to 2018. Furthermore, we extracted more related documents from Google Scholar and reference list. Table 1 shows the search terms used for the databases.

**Table 1. Search terms on Scopus and IEEE Explore (1963 to 2018)**

| # | Searches |
|---|---|
| 1 | Artificial Intelligence |
| 2 | AI |
| 3 | Programming Language |
| 4 | 1 OR 2 |
| 5 | 3 AND 4 |

## 2.3 Selection Criteria

We searched for studies that are related to AI Programming Language. Specifically, we included studies that revealed the features or capabilities, limitations and applications of AI programming languages. We excluded reviews, viewpoints or editorials. We also reviewed studies that were written in English only.

## 2.4 Case Definitions

For this research, we signposted our definition of AI based on the definition put forward by Nils Nilsson [33] in his book, "Artificial intelligence is that activity devoted to making machines intelligent, and intelligence is that quality that enables an entity to function appropriately and with foresight in its environment". This definition would guide our final selection of retained studies in subsequent sections.

## 2.5 Quality Criteria

We ensured that for each of the full text accessed, the AI programming language was explicitly defined and meet the case definition as stated in 2.3. Studies with ambiguous details of the features or capabilities, limitations and applications of AI were excluded from this review.

## 2.6 Data Extraction

Data extraction was conducted by the following reviewers; TJ, AA, FA and OA. The selection of relevant articles was performed by all reviewers. Any disagreement between reviewers on our selection of studies was resolved by the fifth and sixth reviewer EA and JB. From each study, we extracted data on programming language, version, year of execution, development team, capabilities or features, limitations and applications. This paper seeks to present the evolution of AI programming languages, as such, the retained studies are presented in a chronological order and grouped based on the type of the AI programming language.

# 3 Results

## 3.4 Systematic Search

Our search returned 6,253 documents that relates to AI programming languages; Scopus (5818) and IEEE Xplore (435). Furthermore, a thorough manual search was performed on Google scholar and returned 312 documents. Hence, a total of 6,565 documents were prepared for analysis. Afterwards, 432 duplicated documents were removed. Of the 6,133 documents remaining, 5,947 documents were excluded based on the following criteria; studies that do not relate to Artificial Intelligence programming languages and studies that are reviews, books, book chapters, report, notes, short survey, letter, viewpoints and editorials. A total of 186 full texts and abstracts were assessed for eligibility, thereafter a total of 69 studies were retained for qualitative and quantitative synthesis. The highlighted search procedure is graphically represented in Figure. 1.

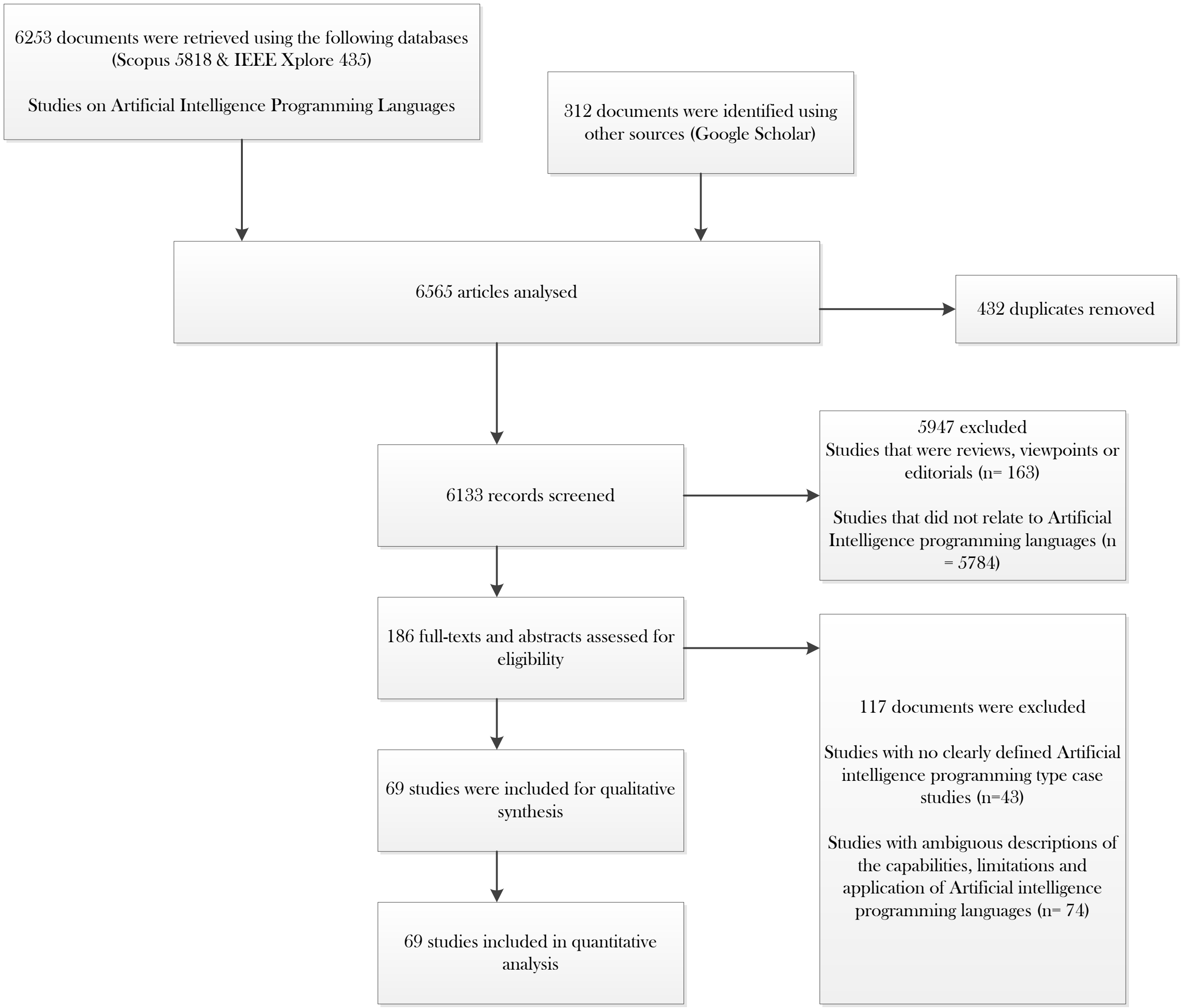

**Figure 1. PRIMA flowchart for selected studies**

## 3.5 Study Characteristics

The bulk of the retained studies on Artificial Intelligence programming languages can be grouped under the main categories LISP, PROLOG. Two studies in 1986 and 1987 [34-35] respectively dealt with different variants of Prolog namely TC-Prolog and FProlog. In recent years, we see documentations on new Artificial Intelligence programing language like Logic and Objected-oriented programming language (LOOP) documented by [36]in 2001 which extended Prolog logic programming language with object oriented features while [37] discuss a connection architecture between Prolog and JAVA, Archlog documented in 2006 [38] which can produce high-performance designs without detailed knowledge of hardware development and a framework for designing multiprocessor architectures; Epistemic Ontology Language with Constraints (EOLC) which is used for specifying the epistemic ontology for heterogeneous verification, documented in 2007 by [39]; McKinley described Python in his 2006 paper [40], [40-43] in 2012,2015,2016 and 2017 respectively discussed the use of C++ in Artificial intelligence programming languages while [28, 42, 44-47] discussed the use of JAVA in Artificial intelligence programming languages.

## 3.6 Successes, Challenges And Limitations

### 3.6.1 Successes of Artificial Intelligence

Optimization of AI language has enabled a revolutionary change across many sectors globally. For instance, AI has been deployed across different fields like finance, health, engineering and education and The proliferation of AI languages across these sectors has helped birth new tech businesses [48]. In Education and Research, AI has brought about significant improvement in the quality of delivery of educational resources across the globe, expert systems have been created to provide seamless learning content to students and researchers across boarders [49]. In the health sector, AI has been used to assist in automated data management [50], development of artificial neural network to assist in rapid patient care [51]. The advancements in AI has provided further dynamics of analysis of face or object recognition techniques for audiovisual. Also in music evolution in composing human like notes [52].The invasion of cognitive problem-solving skill has brought innovative ideas that is answering engineering questions[48].

### 3.6.2 Challenges and Limitations of Artificial Intelligence

The main aim of artificial intelligence is to build an intelligent machine that will make life easier for human beings. The machine should be able to think like humans and some intelligence traits added to it. The programmers want to build some emotional quotient into the machines. Expert development is one of the major problems of AI programming language. LISP; a functional language was created as a mathematical notation for computer programmer was developed for lambda calculus which is not part of undergraduate curriculum in higher institutions. This makes it more tedious for beginners to master LISP compare to other object -oriented language like JAVA. The expert community and library capacity is limited due to this difficulty [49, 53-54].

Since AI involves building a machine that is intelligent like human beings, it must also face some challenges like humans. Identifying some of these challenges will minimize the associated risks and at the same time make sure that we take full advantage of this technology

Most researchers believe and agree that a super intelligence AI is unlikely to showcase human emotions like love or hate and that therefore, it cannot become intentionally benevolent or malevolent. However, the most likely scenario where it can pose a threat to the society is via autonomous weapons. These are weapons that AI systems are programmed to use to kill. If in the hands of the wrong person, these weapons could easily cause mass casualties. This could even lead to an AI war that would also result in mass causalities [55].

Legal challenges related to AI's application in the financial industry could be related to the consequences of erroneous algorithms and data governance. Erroneous algorithms, due to the lack of appropriate data, can leave a big dent in the profits of an organization by making incorrect and perhaps detrimental predictions. Poor data governance can result in data breaches where customers' PII (personal identifiable information) that acts as a feedstock to an algorithm may get into the hands of hackers and can cause legal challenges for the organization [56].

Table 2. Extracted Studies on LISP Artificial Intelligence programming languages

| Programming Language | Version | Authors | Year of execution | Development team | Capabilities/features | Limitations | Applications |
|---|---|---|---|---|---|---|---|
| LISP | Lisp 1.5 | John, McCarthy [57-61] | 1962 or 1985 (Not sure) | Research Laboratory of Electronics, Massachusetts Institute of Technology, U.S. Army, U.S. Navy and U.S. Air Force | Lisp is a mathematical language that uses Symbolic data processing (S-expressions). These S-expressions are stored in a structured list.<br><br>It performs computational analysis on sequential programs, it has simple internal structure and compatible with other systems. | Ultra slow numerical computation and lack of better representation of block of registers.<br><br>It has a higher overhead when compared with other conventional programming languages. This has narrowed its use in AI.<br><br>It is limited with respect to processing and memory requirements. Its size, unwieldiness, "kitchen sink" design strategy, and general ADAfication is disliked by most of its critics.<br><br>Slower compare to C++ | Used in differential and integral calculus, electrical circuit theory, mathematical logic, game playing and other fields of artificial intelligence.<br><br>A vastly useful tool for the programming of many Expert Systems.<br><br>It is programming language of choice of most AI researchers<br><br>It has been applied in Musical composition and processing Foundational language for all programming languages. |
| | | Dixon, Brent W [62] | 1986 | EG&G Idaho Inc, Idaho Falls, ID,, USA. | Able to rapidly analyze large trees by efficiently utilizing a list-based tree structure, search space and rule-based | | |
| | | Hawkinson, Lowell [3] | 1986 | LISP Machine Inc, Cambridge, MA, USA | More powerful for providing symbolic computing compared with conventional programming methods. | | |
| | | S. Kaisler [63] | 1986 | | Very flexible programming language compared with most others because it is built around a kernel of mathematical principles. | | |
| | | Hahne, K.<br>Moeckel, P.<br>Thuerlings, K. J.[64] | 1988 | Siemens AG, Munich, West Ger, Siemens AG, Munich, West Ger | LISP, a Programming Language and Its Computational Models | | |
| | | Rahn, John[65] | 1990 | The University of Washington | Its features includes, a data structure for music | | |
| | | | | | representation, Lispfront (a language for composing music), it is object-oriented and the ability to output Music4P score data | | |
| | | Takeuchi, I.[66] | 2002 | Computer Science Department, University of Electro-Communications, Chofu, 182-8585, Japan | Lisp has potentials for systems programming and is suitable for writing operating systems. It also possess capabilities for heterogeneous parallel computing | | |
| | | Ellis IC 2012 [41] | 2012 | | Coding is fast and efficient due compilers Automatic garbage collections was invented for lisp language | | |
| | | Karnick, H. (2014).[67] | 2014 | | | | |
| | | Rajaraman, V. (2014)[68] | 2014 | | | | |
| | | Priestley, M. (2017)[53] | | | | | |

**Table 3. Extracted Studies on PROLOG Artificial Intelligence programming language**

| Programming Language | Version | Authors | Year of execution | Development team | Capabilities/features | Limitations | Applications |
|---|---|---|---|---|---|---|---|
| PROLOG | | K. Clark and S. Gregory[2] | 1986 | Dept. of Computing, Imperial College of Science and Technology, University of London | Performs efficient implementation by incorporating parallel mode of computation | The coding of the knowledge and the organization and modularization requires a lot of creativity. Some examples include fact/consequent representation problem and existential | Suitably adapts to applications requiring implementation in logic programming environments. Applied in |
| | | Wilkerson, Ralph W.[69] | 1986 | Univ of Missouri, Rolla, MO, USA, Univ of Missouri, | | | |

<table>
<tr><td rowspan="10"></td><td></td><td></td><td></td><td>Rolla, MO, USA</td><td></td><td rowspan="10">quantification problems.<br><br>One of its drawbacks is the lack of natural mechanism to tackle the issue of uncertainties since by default it is designed to be a two-valued logic programming language.<br><br>Prolog by default provides limited support for real-life knowledge engineering.<br><br>In order to improve performance in Prolog applications, the performance of the microarchitecture of the uniprocessor engine needs to be developed.<br>Prolog side-effects are typical necessary evil.<br><br>Convenient logic program properties no longer hold, and upsetting practical problems sprout, namely the debugging of programs with side-effects is harder.<br><br>By default Prolog does not take into account the dynamic nature of agents such as</td><td rowspan="10">Octree solid modeling applications.<br><br>Relational Database applications, natural language processing, theorem proving, automated reasoning.<br><br>Used for establishing expert systems in specific research activities.<br><br><br>Used in logic-circuit model building, expert systems, AI and natural-language interfacing.<br><br>Used in clinics for detection and classification of QRSs in Electrocardiography (ECG).</td></tr>
<tr><td></td><td>Wilson, W. G.[70]</td><td>1986</td><td></td><td>PROLOG for applications programming</td></tr>
<tr><td></td><td>Guerrieri, Ernesto<br>Grover, Vinod[71]</td><td>1986</td><td>SofTech Inc, Waltham, MA, USA,</td><td>Suitable support for programming to modeling with Octrees</td></tr>
<tr><td></td><td>Weeks, J.<br>Berghel, H.[4]</td><td>1986</td><td>Department of Computer Science, University of Nebraska, Lincoln, NE 68588, United States</td><td>Support for relational databases, natural language processing and automated reasoning</td></tr>
<tr><td></td><td>Herther, Nancy K.[72]</td><td>1986</td><td>Univ of Minnesota Humanities/Social, Science Libraries, Minneapolis, MN,, USA</td><td>PROLOG to the future: a glimpse of things to come in artificial intelligence</td></tr>
<tr><td></td><td>Zewari, S. W.<br>Zugel, J. M.[73]</td><td>1986</td><td>Virginia Polytechnic Inst &, State Univ, Blacksburg, VA, USA</td><td>It supports symbolic programming, therefore, by manipulating trigonometric relations and identities, it is able to derive useful kinematic equations of open kinematic chains</td></tr>
<tr><td></td><td>Broberg, Harold L.[74]</td><td>1987</td><td>Indiana-Purdue Univ, Ft. Wayne, IN,, USA, Indiana-Purdue Univ, Ft. Wayne, IN, USA</td><td>It is easy to learn and easily adaptable</td></tr>
<tr><td></td><td>Butrick, Richard[75]</td><td>1987</td><td>Ohio Univ, Athens, OH, USA</td><td></td></tr>
<tr><td></td><td>Giakoumakis, E.<br>Papakonstantinou, G. [50]</td><td>1987</td><td>Natl Technical Univ of Athens, Athens, Greece</td><td>It's facts, rules, questions and inference features are used in recognition tasks in clinical analysis</td></tr>
<tr><td></td><td>Martin, T. P.</td><td>1987</td><td>Information</td><td></td></tr>
</table>

| | | | | | | | |
|---|---|---|---|---|---|---|---|
| | | Baldwin, J. F.<br>Pilsworth, B. W.[35] | | Technology Research Centre, Department of Engineering Mathematics, University of Bristol, Bristol, BS8 1TR, United Kingdom | | knowledge acquisition and action execution.<br><br>This poses a problem because the agent might work in a dynamic environment where unexpected things can happen.<br><br>Slower compare to C++ | Forth-based Prolog is used as basis of the expert system component of an astronaut interface for a series of Spacelab experiments<br><br>It is a well adapted technology for expert systems design<br>Used in real-time expert system to develop astronaut interface for a series of Spacelab vestibular experiments.<br><br>Used in the probing of a facility's defenses and to find potential attack paths that meet designated search criteria.<br><br>Used in electric utility |
| | | Odette, L. L.<br>Paloski, W. H.[76] | 1987 | Applied Expert Systems Inc,, Cambridge, MA, USA, | Forth-based Prolog possess the features for symbolic reasoning | | |
| | | Orci, Istvan P.<br>Knudsen, Erik[77] | 1987 | Univ of Stockholm, Stockholm | Knowledge Engineering Library (KEL) written in Prolog, expands Prolog's knowledge processing functionalities by supporting rule-based knowledge representation and approximate reasoning.<br><br>KEL is capable of reasoning and dealing with both uncertain facts and uncertain rules in a formal way by employing possibilistic logic and fuzzy set theory as its logical basis. | | |
| | | Paloski, William H.<br>Odette, Louis L.<br>Krever, Alfred J.<br>West, Allison K.[78] | 1987 | KRUG Int, Houston, TX, USA, | Forth-based Prolog includes a predicate that can be used to execute Forth definitions. It also has the facility for rule based clauses and a procedure base containing Prolog goals that provides support for coding in Forth. | | |
| | | Pelin, Alex<br>Millar, Robert[79] | 1987 | Florida Int Univ, Miami, FL, USA, Florida Int Univ, Miami, FL, USA | PROLOG and automatic program generation from specifications | | applications, for volt/VAR dispatch and to increase the capability of electric energy management centers to successfully monitor power system operation and promptly respond to emergencies.<br><br>Used in representation of a nursing knowledge base.<br><br>Finds application in HICOM communication system language of choice for knowledge-processing systems suitable for implementing an inference engine for agents. |
| | | Zimmerman, B. D.<br>Densley, P. J.<br>Carlson, R. L.[80] | 1987 | Westinghouse Hanford Co, Richland,, WA, USA, | | | |
| | | Zimmerman, Jennifer [81] | 1987 | Quintus Computer Systems, Mountain, View, CA, USA | It operates on the principles of logic using the same inference methods that define symbolic logic (predicate calculus) that human's use in reasoning. It has the ability to manipulate complex data structures and flexibly represent real-world knowledge greatly speeds development time | | |
| | | Anon[82] | 1988 | | | | |
| | | Ozbolt, Judy G.<br>Swain, Mary Ann P. [83] | 1988 | Univ of Virginia, Sch of Nursing,, Charlottesville, VA, USA | Provides support for graphical illustrations, natural language text, description of major knowledge domains and depictions of the relationships within and among them. | | |
| | | Beer, Armin<br>Manz, Joachim [84] | 1989 | Siemens AG Osterreich, Austria | It is user-friendly and many user tasks can be carried out faster and more economically than with other customary languages such as COBOL or PASCAL. | | |
| | | Buzzi, R. [85] | 1989 | Comparative Physiology and | Allows definition and characterization of groups | | |
| | | | | Behavioral Biology Laboratory, Federal Institute of Technology, Zurich, Switzerland | of subjects and single objects. Classifications performed using PROLOG compares very well with the methods of logistic regression and with discriminant analysis. | | Finds applications in insulin pump systems in hospitals for treatment of diabetic patients.<br><br>Widely used in artificial intelligence research Construction of expert systems.<br><br>Applied in the interpretation of IR spectra |
| | | Colmerauer, A. [86] | 1989 | Faculté des Sciences de Luminy, Unité de recherche associée au Cnrs 816, Case 901, 70, route Léon Lachamp, F-13288 Marseille Cedex 9 | Prolog III programming language expands Prolog functionalities by redefining the fundamental process at its heart, integrates into this mechanism, refined processing of trees and lists, greater number processing, and processing of complete propositional calculus. | | |
| | | Patt, Y. N. [87] | 1989 | Computer Science Division, University of California, Berkeley, CA, United States | | | |
| | | Pereira, L. M.<br>Calejo, M. [88] | 1989 | Logic Programming and Artificial Intelligence Group, Universidade Nova de Lisboa (UNL), Monte da Caparica, Portugal | | | |
| | | Hayashi, H.<br>Cho, K.<br>Ohsuga, A.[89] | 2002 | Computer and Network Systems Laboratory, Corporate Research and Development Center, TOSHIBA Corporation | Prolog-like procedures can be developed to resolve issues of implementing inference engine for dynamic agents | | |
| | | Ellis, I. C. and A. | | Prolog has a built-in | Probabilistic measuring, | | |
| | | Agah[41] | | list handling essential in representing tree-based data structures and pattern matching .it can also backtrack automatically | robotic domain and biological domain<br>The performance Rate is highly improved when connected to Java | | |
| | | Ostermager [37] | 2014 | | | | |
| | | De Raedt, L. [90] | 2015 | | | | |
| | | Jaśkiewicz, G. [91] | 2016 | | | | |
| | | Nickles, M. [92] | 2016 | | | | |
| | | Morozov, A. A., et al [93] | 2017 | | | | |
| | | Zhang, S., et al. (2017)[94] | | | | | |
| PROLOG | TC-PROLOG | Futo, Ivan<br>Papp, Imre[34] | 1986 | Computer Research Inst, Budapest, Hung, | Ability to strategically search for the right dosage of drug by leveraging on a combination of pharmacokinetic/pharmacodynamic models and logical decision rules | | |
| PROLOG | FPROLOG (Fuzzy prolog) | Martin, T. P.<br>Baldwin, J. F.<br>Pilsworth, B. W.[35] | 1987 | Information Technology Research Centre, Department of Engineering Mathematics, University of Bristol, Bristol, BS8 1TR, United Kingdom | FPROLOG Builds generality and flexibility over conventional Prolog by empowering the language with breadth-first and depth-first search capabilities. | | |
| PROLOG, LISP | | Postma, G. J.<br>Vandeginste, B. G. M.<br>van Halen, C. J. G.<br>Kateman, G.[95] | 1987 | | They were designed to primarily manipulate symbols rather than numbers. It, therefore, possess toolkits that consist of various knowledge representation methods and inference engines | | |
| | | | | | (Useful to get full paper) | | |
| PROLOG, LISP | | Postma, G. J.<br>Vandeginste, B. G. M.<br>van Halen, C. J. G.<br>Kateman, G.[96] | 1987 | Department of Analytical Chemistry, Faculty of Science, University of Nijmegen, Toernooiveld, Nijmegen, The Netherlands | Implementation of a teaching program for IR spectrometry in Lisp and Prolog<br>(Useful to get full paper) | | |
| PROLOG,LISP | | Falk, Howard[97] | 1988 | Computer Design, United States | | Compared with coding in conventional languages such as C or Ada, code executions in PROLOG or LISP is slower. In addition to this, their memory demands are extensive. | |

**Table 4. Extracted Studies on LOOP Artificial Intelligence programming language**

| Programming Language | Version | Authors | Year of execution | Development team | Capabilities/features | Limitations | Applications |
|---|---|---|---|---|---|---|---|
| LOOP (logic and object-oriented programming language) | | A. Suciu<br>K. Pusztai<br>T. Muresan<br>Z. Simon[36] | 2001 | Dept. of Comput. Sci., Tech. Univ. of Cluj-Napoca, Romania | LOOP extends Prolog logic programming paradigm with object-oriented features | Prolog lacks mechanisms for structuring knowledge (program clauses) | Used in LP (Linear Programming)-based AI applications |

**Table 5. Extracted Studies on ARCHLOG Artificial Intelligence programming language**

| Programming Language | Version | Authors | Year of execution | Development team | Capabilities/features | Limitations | Applications |
|---|---|---|---|---|---|---|---|
| Archlog | | Fidjeland, A. Luk, W.[38] | 2006 | Imperial College London, 180 Queen's Gate, London SW7 2AZ, United Kingdom | Can produce high-performance designs without detailed knowledge of hardware development and a framework for designing multiprocessor architectures | | Finds application in machine learning and cognitive robotics |

**Table 6. Extracted Studies on EOLC Artificial Intelligence programming language**

| Programming Language | Version | Authors | Year of execution | Development team | Capabilities/features | Limitations | Applications |
|---|---|---|---|---|---|---|---|
| EOLC (Epistemic Ontology Language with Constraints) | | R. Kumar B. H. Krogh[39] | 2007 | Department of Electrical and Computer Engineering, Carnegie Mellon University, Pittsburgh, PA 15213, USA. | Used for specifying the epistemic ontology for heterogeneous verification | | Finds application in redundant flight guidance system and in the heterogeneous Verification of Embedded Control Systems |

**Table 7. Extracted Studies on PYTHON Artificial Intelligence programming languages**

| Programming Language | Version | Authors | Year of execution | Development team | Capabilities/features | Limitations | Applications |
|---|---|---|---|---|---|---|---|
| PYTHON | | McKinley, K. S. (2016)[40] | 1991 | | Python has improved over a short period of time compare to Java and C++<br><br>The efficiency of the programmer is highly improved because of its support for object-oriented design, functional and procedural styles of programming.<br><br>It has high level syntax<br><br>Algorithm can be tested without implementation. | Is not good for mobile computing because of its weak Language for mobile computing<br><br>The execution is slow in artificial intelligence development due to the fact that it works with the help of an interpreter unlike C++ and Java | |

Table 8. Extracted Studies on C++ Artificial Intelligence programming languages

| Programming Language | Version | Authors | Year of execution | Development team | Capabilities/features | Limitations | Applications |
|---|---|---|---|---|---|---|---|
| C++ | | Park, J., et al. (2017)[43] | 2017 | | C++ can organize data because it is a multi-paradigm programming that supports object-oriented principles.<br><br>It has high level of abstraction which makes it good for solving complex problem in AI | C++ can organize data because it is a multi-paradigm programming that supports object-oriented principles.<br>It has high level of abstraction which makes it good for solving complex problem in AI | Used in virtual robot simulation and synthesis, grasp synthesis, 3D drawing.<br>Any kind of data can be modelled and simulated easily in AI |
| | | McKinley, K. S. (2016)[40] | 2016 | | | | |
| | | Kurniawan, A., et al. (2015)[42] | 2015 | | | | |
| | | Ellis, I. C. and A. Agah [41] | | | | | |

Table 9. Extracted Studies on JAVA Artificial Intelligence programming languages

| Programming Language | Version | Authors | Year of execution | Development team | Capabilities/features | Limitations | Applications |
|---|---|---|---|---|---|---|---|
| JAVA | | Ostermayer 2014[37] | | | Java is portable<br>Implementation on different platforms easy because of virtual machine technology<br>Algorithms coding is very easy | Java's Response time is more and less execution speed this makes it slower than C++ | It enables the Automatic Speech Recognition (ASR) systems Improved performance when linked with prolog |
| | | Babu, A. A., et al. (2015)[44] | | | | | |
| | | Kurniawan, A., et al. (2015)[42] | | | | | |
| | | Mittal, H. and S. D. Mandalika (2015)[45] | | | | | |
| | | McKinley, K. S. (2016)[40] | | | | | |
| | | Garg, S. and S. Kumar (2017)[46] | | | | | |
| | | Raff, E. (2017).[47] | | | | | |

Table 10. Extracted Studies on HASKELL Artificial Intelligence programming languages

| Programming Language | Version | Authors | Year of execution | Development team | Capabilities/features | Limitations | Applications |
|---|---|---|---|---|---|---|---|
| HASKELL | | Magalhães, J. P. and A. Löh (2015)[98] | 1990 | | the algorithms can be gotten in cabal<br>Can be simulated on CPU cloud ,bytecode complier and CUDA binding. | | Uses algebraic datatype to reduce code duplication in generic programming. |

Table 11. Extracted Studies on ADA Artificial Intelligence programming languages

| Programming Language | Version | Authors | Year of execution | Development team | Capabilities/features | Limitations | Applications |
|---|---|---|---|---|---|---|---|
| ADA | | Hattori,Kushima,Wasano[99] | 1985 | | High performance and maintainability, list processing facility | | Development of large scale AI software |

## 4 Discussion

This study was conducted to give a clear understanding and relevance of AI programming language from conception till date. The study confirms that John McCarthy pioneered the concept of AI which has evolved into the implementation of other high level programming languages such as LISP, Prolog, etc. Our findings suggest that the field of AI; a multi-disciplinary field, can be highly enriched if further research is geared towards the development of more syntax and semantic interaction that could produce a robust language understanding systems.

We discuss the answers to our research questions in the following sub-sections.

### *4.1 What is the Prevalence of AI Programming Language Publications Since 1963?*

AI programming languages have seen significant research interest since 1963. The research area has seen different peaks and valleys in research outputs. Between 1980 and 1986, there was a steep increase of 8 to 242 documents in SCOPUS, a significant reduction between 1987 and 2003 before increasing to 278 in 2004, between 2005 and 2014, a dip occurred in research output and increased back to 275 in 2015.

### *4.2 Which of the AI Programming Languages have Received more Attention with Respect to the Volume of Research Publication Being Produced?*

The volume of research outputs focusing on the PROLOG programming language has been rather significant. From Table 3, 34 research publications were reviewed for Prolog in this systematic review, 15 documents were reviewed for the LISP programming language while the remaining 20 documents were reviewed for Logic and Object oriented programming (LOOP), ARCHLOG, Epistemic Ontology Language with Constraints (EOLC), Phyton, C++, ADA and JAVA programming languages making it a total of 69 reviewed articles.

### *4.3 What are the Predominant AI Programming Languages on which Recent AI Software's Rely?*

A number of AI libraries have been written based on some of the reviewed foundational AI programming languages; some of these programs are listed in Table 12. As illustrated in the Table, C/C++ has enjoyed the highest level of patronage (N=7) by developers of modern AI libraries. This is closely followed by Python (N=6). The reasons for the high adoption of these languages may be attributed to their unique strengths compared with other reviewed languages as earlier illustrated in Tables 7 and 8.

**Table 12. Details of Modern AI Libraries and their Foundational AI Programming Languages**

| S/N | Modern AI Libraries | Initial Release data | Original Author | AI programming language platform |
|---|---|---|---|---|
| 1 | ALICE | 1995 | Joseph Weizenbaum | Java |
| 2 | OpenNN | 2003 | International Center for Numerical Methods in Engineering(CIMNE) | C++ |
| 3 | OpenCog | | | C++, Python |
| 4 | TensorFlow | 2015 | Google Brain Team | Phyton, C++, CUDA |
| 5 | Siri | 2011 | Apple | Objective C |
| 6 | Neural Designer | | Arteinics | C++ |
| 7 | Keras | 2015 | | Python |
| 8 | Scikit-learn | 2007 | David Cournapeau | Python, Cython, C, C++ |
| 9 | Pandas | 2017 | Wes McKinney | Python |
| 10 | SciPy | 2017 | Travis Oliphant, Pearu Peterson, Eric Jones | Python Fortran, C, C++ |

### *4.4 What are the Characteristics of AI Programming Languages which make them Suitable or Unsuitable for use across Pplatforms?*

One of the most desirable qualities of AI programming languages is speed. PROLOG and LISP are slower compared to C/C++ and Python. An additional ability to support object oriented principles is desirable. Sometimes it may be necessary to trade between quick response time and execution speed. Other desirable qualities are portability and ease of coding. The combination of C/C++ with Python in developing most of the trending AI libraries (such as TensorFlow and Keras) as illustrated in Table 12 may be born out of the need to complement the weakness of one language with the strength of the other and vice-versa.

## 5. Conclusion

In each of the sections, we have covered different areas of AI that carry similar coverage which resulted into the applications side and a lot more on the evolution of the AI languages. However, great progress has been made in the field which has grown over the last 55 years and practically helped to solve many potential human problems concerning machine vision, game-playing, and automatic language processing among others. This has even stretched beyond the scope of the originators of AI like Turing and McCarthy ever envisaged. More researches were geared towards fine tuning the programming languages not only being functional but also having a trace of embedded logic capabilities. Conclusively, as a science of intelligence, AI can be legitimately looked upon in the field of psychology because of its relevance to cognitive psychology.

## Acknowledgement

The authors wish to acknowledge the Covenant University Center for Research, Innovation and Discovery (CUCRID) for providing fund towards the publication of this study.